\documentclass[12pt]{article}
\usepackage{arxiv}
\usepackage{natbib}

\usepackage[utf8]{inputenc} 
\usepackage[T1]{fontenc}    
\usepackage{hyperref}       
\usepackage{url}            
\usepackage{booktabs}       
\usepackage{amsfonts}       
\usepackage{nicefrac}       
\usepackage{microtype}      
\usepackage{graphicx}
\usepackage{xcolor}

\usepackage{tablefootnote}

\usepackage{subcaption}
\usepackage{caption} 
\usepackage{amsmath,amssymb}
\newcommand{\x}{\mathbf{x}}
\newcommand{\z}{\mathbf{z}}
\newcommand{\w}{\mathbf{w}}

\renewcommand{\u}{\mathbf{u}}
\newcommand{\y}{\mathbf{y}}

\newcommand{\f}{\mathbf{f}}

\newcommand{\K}{\mathbf{K}}

\newcommand{\LL}{\mathbf{L}}
\newcommand{\E}{\mathbf{E}}
\newcommand{\bt}{{\boldsymbol{\theta}}}

\newcommand{\R}{\mathbb{R}}

\newcommand{\N}{\mathcal{N}}
\newcommand{\GP}{\mathcal{GP}}

\newcommand{\W}{\mathbf{W}}

\newcommand{\bmu}{\boldsymbol{\mu}}
\newcommand{\s}{{\sigma}}
\renewcommand{\S}{{\boldsymbol{S}}}

\newcommand{\bb}{\mathbf{b}}

\newcommand{\m}{\boldsymbol{m}}
\newcommand{\h}{\mathbf{h}}

\renewcommand{\vec}[1]{\boldsymbol{#1}}
\newcommand{\mat}[1]{\boldsymbol{#1}}

\DeclareMathOperator{\cov}{\textbf{cov}}

\DeclareMathOperator{\logit}{logit}

\DeclareMathOperator{\KLd}{KL}
\newcommand{\KL}[2]{{\KLd({#1} || {#2})}}

\begin{document}

\title{Neural Non-Stationary Spectral Kernel}

\author{%
Sami Remes \\
\texttt{sami.remes@aalto.fi} \\
$^1$Department of Computer Science, \\
Aalto University, Espoo, Finland \\
$^2$Silo AI, Helsinki, Finland \\
\And
Markus Heinonen \\
\texttt{markus.o.heinonen@aalto.fi} \\
Department of Computer Science, \\
Aalto University, Espoo, Finland \\
\And 
Samuel Kaski \\
\texttt{samuel.kaski@aalto.fi} \\
Department of Computer Science, \\
Aalto University, Espoo, Finland}

\maketitle

\begin{abstract}
Standard kernels such as Matérn or RBF kernels only encode simple monotonic dependencies within the input space. Spectral mixture kernels have been proposed as general-purpose, flexible kernels for learning and discovering more complicated patterns in the data.
Spectral mixture kernels have recently been generalized into non-stationary kernels by replacing the mixture weights, frequency means and variances by input-dependent functions.
These functions have also been modelled as Gaussian processes on their own. 
In this paper we propose modelling the hyperparameter functions with neural networks, and provide an experimental comparison between the stationary spectral mixture and the two non-stationary spectral mixtures.
Scalable Gaussian process inference is implemented within the sparse variational framework for all the kernels considered. We show that the neural variant of the kernel is able to achieve the best performance, among alternatives, on several benchmark datasets.
\end{abstract}
\keywords{Gaussian process, neural networks, non-stationary kernel, spectral mixture kernel}

\section{Introduction}

Gaussian processes are a Bayesian non-parametric method often applied for non-linear function approximation in tasks such as regression or classification \citep{rasmussen2006}. They essentially define a distribution over functions, and their performance depends heavily on the chosen kernel function that encodes the prior beliefs about the properties of the function, effectively constraining the behaviour of the function. 
Many popular kernels, such as the Gaussian and the Mat\'ern kernels, lead to smooth neighborhood-dominated interpolation that is oblivious of long-range and periodic dependencies within the input space, and can not adapt the similarity metric in different parts of the input space due to being stationary.

Two important properties of kernels are \emph{stationarity} and \emph{monotony}. A stationary kernel $k(x,x') = k(x+a,x'+a)$ is a function only of the distance $\tau = x-x'$ and not directly the value of $x$, encoding an identical similarity across the whole input space. A monotonic kernel decreases over the distance $\tau$, being unable to consider periodicity or long-range connections. Kernels that are both stationary and monotonic, such as the Gaussian and Mat\'ern kernels, can learn neither input-dependent function dynamics nor long-range correlations within the input space. Non-monotonic and non-stationary functions are common in realistic signal processing \citep{rioul1991}, time series analysis \citep{huang1998}, bioinformatics \citep{Grzegorczyk2008,robinson2009}, and in geostatistics applications \citep{higdon1999,huang2008}.

Recently, \citet{remes2017} proposed an expressive kernel family that is both non-stationary and non-monotonic, and hence can infer long-range or periodic relations in an input-dependent manner, as a generalisation of the spectral mixture kernel \citep{wilson2013}.
The input-dependent parameterisation was achieved by modelling the frequency parameters themselves as Gaussian processes.
In this paper we propose a more flexible, deep neural network parameterization instead. This achieves both better performance in terms of log-likelihood and errors, as well as being faster in wall-time per iteration, during optimisation of the kernel parameters.
We show and compare the expressivity and the performance of the proposed kernel with experiments on time series from two solar activity datasets as well as motion capture data.

Several authors have explored kernels that are either non-monotonic or non-stationary. A non-monotonic kernel can reveal informative dependencies over the input space by connecting distant points due to periodic or other effects. Standard periodic kernels \citep{rasmussen2006} exhibit non-monotonic properties, but assume that the periodicity extends infinitely. Non-monotonic kernels have also been derived from the Fourier decomposition of kernels \citep{lazaro2010sparse,sinha2016,wilson2013}, which renders them inherently stationary, however.

Non-stationary kernels, on the other hand, are often based on generalising monotonic base kernels, such as the Gaussian or Mat\'ern family of kernels \citep{heinonen2016,paciorek2004}, by partitioning the input space \citep{gramacy2008}, or by input transformations \citep{snoek2014}. 
The work started by \citet{remes2017}, and further developed in this paper, considers generalising the already flexible and non-monotonic, but stationary, spectral mixture kernel \citep{wilson2013}.

\section{Gaussian Processes}

A Gaussian process (GP) defines a distribution over functions $f(x)$, denoted as
\begin{align}
    f(x) \sim \GP(m(x), k(x,x')), 
\end{align}
where the function $m(x)$ defines the prior mean value $\E[f(x)] = m(x)$ and the kernel function $k(x, x')$ denotes the covariance between function values $\cov[ f(x), f(x') ] = k(x, x')$ at points $x$ and $x'$. Per definition, for any collection of inputs, $x_1, \ldots, x_N$, the function values $f_i = f(x_i)$ follow a multivariate normal distribution 
\begin{align}
    \begin{pmatrix}f_1, & \hdots, & f_N\end{pmatrix}^T \sim \N(\m, \K) \,,
\end{align}
where the elements of $\K$ are $\K_{ij} = k(x_i, x_j)$ and $\m$ are $m_i = m(x_i)$.

The key property of Gaussian processes is that they can encode smooth functions by correlating function values of input points that are similar according to the kernel $k(x,x')$. A very commonly used kernel is the squared exponential, also called the radial basis function (RBF), kernel
\begin{align}
    k_{\text{RBF}}(x, x') = \sigma_f^2 \exp\left( -\frac{(x - x')^2}{2\ell^2}\right) \, ,
\end{align}
where the signal variance $\sigma_f$ corresponds to the range scale of the function values, and the length-scale $\ell$ encodes how fast the function values can change with respect to the distance $\tau = x - x'$. The mean function $m(x)$ is often assumed to be a constant $m(x) \equiv 0$, and the data is normalized to zero mean.

\begin{figure}
    \centering
    \includegraphics[width=1\textwidth]{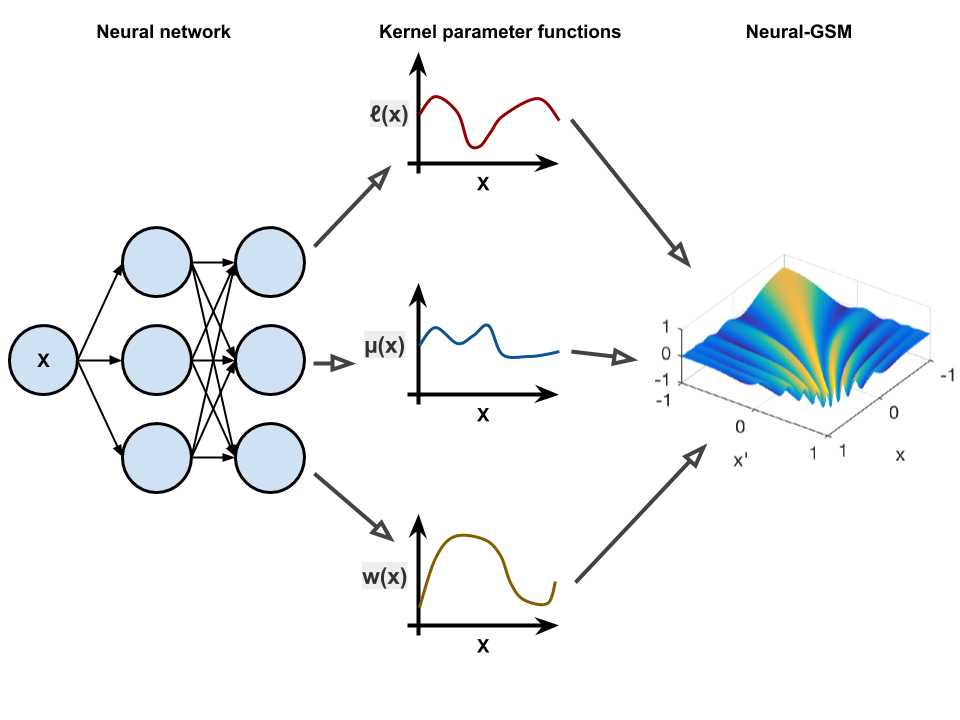}
    \caption{The proposed Neural-GSM kernel applies neural networks (shown on the left) to compute the kernel parameter functions (middle) that are used to compute the kernel function (shown on the right).}
    \label{fig:schematic}
\end{figure}

\section{Spectral and Non-Stationary Kernels}\label{sec:related}

A stationary covariance can with a slight abuse of notation be written as $k(\tau) = k(x-x') = k(x,x')$. 
The main building block for stationary spectral kernel constructions is the Bochner's theorem.
The theorem implies a Fourier dual \citep{wilson2013} for the kernel $k$ and its spectral density $S$:
\begin{align}
k(\tau) &= \int S(s) e^{2\pi i s \tau} ds \\
S(s) &= \int k(\tau) e^{-2\pi i s \tau} d\tau \, . \label{eq:fourier}
\end{align}
This has been exploited to design rich, yet stationary kernel representations \citep{sinha2016,yang2015} and used for large-scale inference \citep{rahimi2008}. 
\citet{lazaro2010sparse} proposed to directly learn the spectral density as a mixture of Dirac delta functions, $S(s) = \sum_i \delta(s-s_i)$ leading to the sparse spectrum (SS) kernel $k_{\text{SS}}(\tau) = \frac{1}{Q} \sum_{i=1}^Q \cos(2 \pi s_i^T \tau)$. 

\citet{wilson2013} derived a stationary spectral mixture (SM) kernel by modelling the univariate spectral density using a mixture of normals 
\begin{align}
    S_{\text{SM}}(s) = \sum_i w_i^2 [\N(s|\mu_i,\sigma_i^2) + \N(s| -\mu_i,\sigma_i^2)] / 2 \,,
\end{align}
corresponding to the kernel function 
\begin{align}
    k_\text{SM}(\tau) = \sum_i w_i^2 \exp(-2\pi^2\sigma_i^2\tau)\cos(2\pi\mu_i\tau) \,.
    \label{eq:sm}
\end{align}
The SM kernel was also extended for multidimensional inputs using Kronecker structure for scalability \citep{wilson2014fast}. 

Kernels derived from the spectral representation are particularly well suited to encoding long-range, non-monotonic or periodic kernels; however, they have been unable to handle non-stationarity before \citet{remes2017}, although \citet{wilson2014thesis} also presented a partly non-stationary SM kernel that has input-dependent mixture weights. \citet{komsamo2015} have also derived general formulations for stationary kernels as well as non-stationary ones, which were based on a more general version of the Bochner's theorem.

Extending standard kernels with input-dependent length-scales \citep{gibbs1997,heinonen2016,paciorek2004,paciorek2006}, input space warpings and transformations \citep{sampson1992,snoek2014,wilson2016deep}, and with local stationarity with products of stationary and non-stationary kernels \citep{genton2001,silverman1957} have been a popular way of constructing non-stationary kernels. The simplest non-stationary kernel is arguably the dot product kernel \citep{rasmussen2006}, which has also been used as a way to assign input-dependent signal variances \citep{tolvanen2014}. These types of non-stationary kernels are a good match for functions with transitions in their dynamics, yet are unsuitable for modelling non-monotonic properties.

\section{Generalised Spectral Mixture (GSM) kernel}

\begin{figure}
\centering
\begin{subfigure}[b]{0.3\textwidth}
    \includegraphics[width=\textwidth]{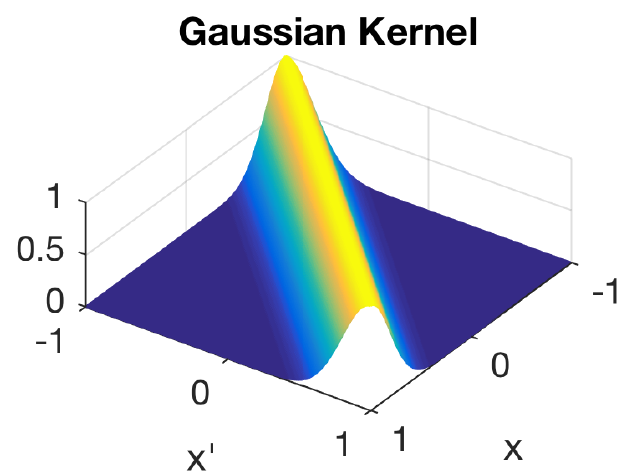}
    \caption{}
\end{subfigure}
~
\begin{subfigure}[b]{0.3\textwidth}
    \includegraphics[width=\textwidth]{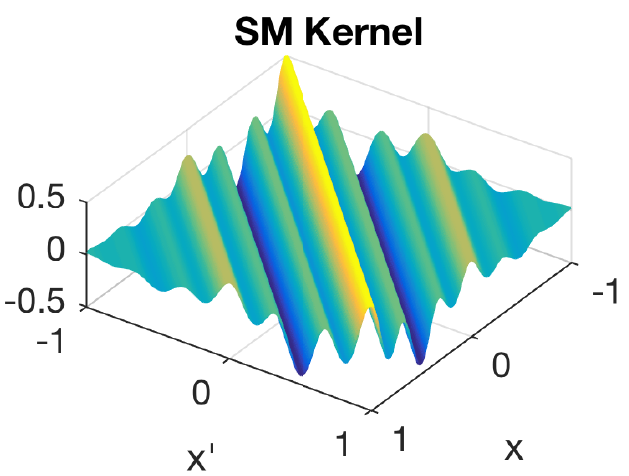}
    \caption{}
\end{subfigure}
~
\begin{subfigure}[b]{0.3\textwidth}
    \includegraphics[width=\textwidth]{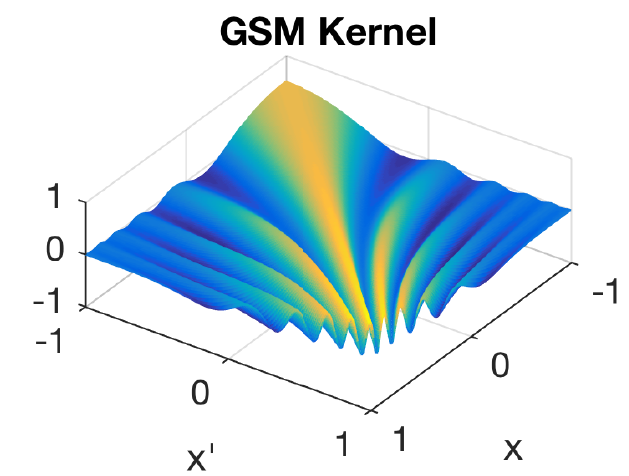}
    \caption{}
\end{subfigure}
\\ \vspace{1em}
\begin{subfigure}[b]{0.3\textwidth}
    \includegraphics[width=\textwidth]{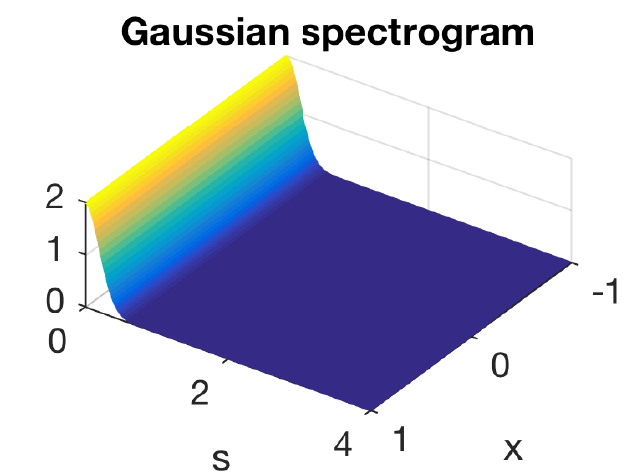}
    \caption{}
\end{subfigure}
~
\begin{subfigure}[b]{0.3\textwidth}
    \includegraphics[width=\textwidth]{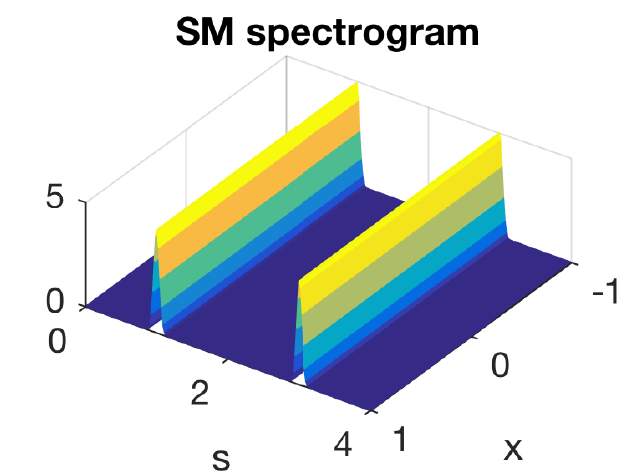}
    \caption{}
\end{subfigure}
~
\begin{subfigure}[b]{0.3\textwidth}
    \includegraphics[width=\textwidth]{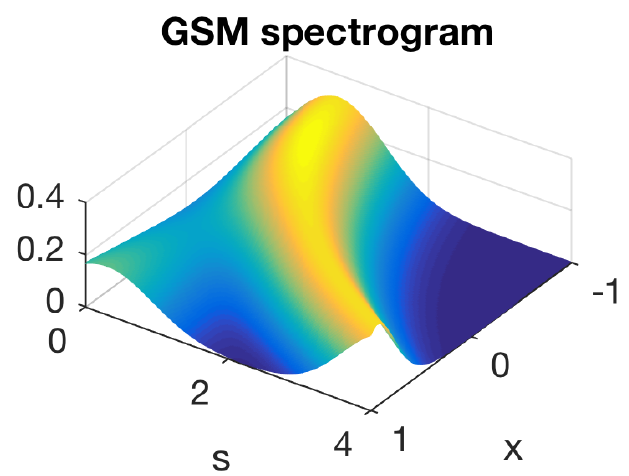}
    \caption{}
\end{subfigure}
\caption{Illustration of the Gaussian (a), SM (b) and GSM (c) kernels on inputs on an interval $[-1, 1]$. Gaussian and SM kernels' corresponding frequency representations (d-e) are constant across input $x$ according to \eqref{eq:fourier}, while for GSM (f) we plot the spectrogram as $S(s, x) = \sum_q w_q^2(x) \N(s \mid \mu_q(x), \ell_q^{-2}(x))$, depicting in this case an increasing frequency component.}
\label{fig:spectrograms}
\end{figure}

\citet{remes2017} presented the generalised spectral mixture (GSM) kernel, that can be seen as a generalisation of the spectral mixture (SM) kernel of \citet{wilson2013}. 
The GSM kernel essentially parameterised the SM kernel hyperparameters by Gaussian processes.
That is, the frequencies $\mu$, length-scales $\ell$ and mixture weights $w$ are given GP priors, that form a smooth spectrogram (in a sense that the frequencies and their amplitudes are functions of $x$, see Figure~\ref{fig:spectrograms} for illustration):
\begin{align}
    \log w_i(x) \sim \GP(0, k_w(x,x')), \\
    \log \ell_i(x) \sim \GP(0, k_\ell(x,x')), \\
    \logit \mu_i(x) \sim \GP(0, k_\mu(x,x')).
\end{align}
Here the log transform is used to ensure the mixture weights $w(x)$ and lengthscales $\ell(x)$ are non-negative, and the logit transform $\logit \mu(x) = \log\frac{\mu}{F_N-\mu}$ limits the learned frequencies between zero and the Nyquist frequency $F_N$, which is defined as half of the sampling rate of the signal (or for non-equispaced signals as the inverse of the smallest time interval between the samples).

The input-dependent lengthscale are accommodated by replacing the exponential part of \eqref{eq:sm} by the Gibbs kernel
\begin{align*}
    k_{\text{Gibbs},i}(x,x') = \sqrt{\frac{2\ell_i(x) \ell_i(x')}{\ell_i(x)^2+\ell_i(x')^2}}\exp\left(-\frac{(x-x')^2}{\ell_i(x)^2+\ell_i(x')^2}\right) \; ,
\end{align*}
which is a non-stationary generalisation of the Gaussian kernel \citep{gibbs1997,heinonen2016,paciorek2004}. 
The non-stationary generalised spectral mixture (GSM) kernel is then given by:
\begin{align}
    k_{\text{GSM}}(x,x') = \sum_{i=1}^Q w_i(x) w_i(x') k_{\text{Gibbs},i}(x,x') \cos(2 \pi (\mu_i(x) x - \mu_i(x') x')) \; . \label{eq:gsm}
\end{align}
The kernel is a product of three PSD terms. The GSM kernel encodes the similarity between two data points based on their combined signal variance $w(x)w(x')$, and the frequency surface based on the frequencies $\mu(x),\mu(x')$ and frequency lengthscales $\ell(x),\ell(x')$ associated with both inputs.
We will refer to this variant of the GSM kernel as GP-GSM in the following.

\section{Parameterizing GSM with Neural Networks}

The latent functions modelled as Gaussian processes pose several issues.
First, we need to be able set the hyperparameters for their kernels, including e.g. the length-scale which determines the smoothness of the function. The needed length-scale value may not be obvious, and needs to be selected by trying many values using e.g. cross-validation approaches which can become very costly computationally. 
Second, the GP features are not exploited to the fullest extent, as the uncertainty implied by the GP interpolation is not utilized, as the kernel values are only based on the mean interpolant. 

In this paper, we propose modelling the latent frequency, variance and length-scale functions using neural networks instead. This is similar conceptually to variational autoencoders \citep{kingma2013auto}, here we apply the neural networks to model parameters of a Gaussian process kernel instead of parameters of a variational distribution. We will refer to this variant of the GSM kernel as Neural-GSM in the rest of this paper. 

A typical, fully-connected feed-forward neural network \citep{goodfellow2016deep} is defined as a composition of simple affine maps (referred to as layers) followed by element-wise applied non-linear transformations (activations). This can be written as
\begin{align}
    \h_1 &= a(\W_1 x + \bb_1) \\
    \y &= a(\W_2 \h_1 + \bb_2)
\end{align}
for a network with just one hidden layer $\h_1$, for a function from $x$ to $\y$. We can also denote this as $\y = \text{NN}(x)$. For the hidden layer activation functions, we employ the scaled exponential linear units (SELU) that have been proposed recently for feed-forward networks \citep{klambauer2017self}. At the top layer, we use the softplus activation
\begin{align}
    \operatorname{softplus}(x) = \log(1+\exp(x))
\end{align}
which ensures that the functions are positive-valued. Now the latent function within the GSM kernel are given as
\begin{align}
    \w(x) = \text{NN}_w(x) \\
    \vec\ell(x) = \text{NN}_\ell(x) \\
    \vec\mu(x) = \text{NN}_\mu(x) \, ,
\end{align}
where all $Q$ elements of each functions are given by the same neural network, that is they share all weights except on the final layer before the softmax activation. Thus the neural network is able to find a common representation of the input data that is useful in determining the amplitudes $\w$, the lengthscales $\vec\ell$ and frequencies $\vec\mu$. This is in contrast to the GP-GSM, where the functions within each mixture component were modelled independently.

\section{Properties of the GSM kernel}\label{sec:properties}

\subsection{Relation to Stationary Spectral Mixture}

We confirm here that the proposed non-stationary GSM kernel reduces to the stationary SM kernel with appropriate parameterisation. We show this identity for univariate inputs for simplicity, with the same result being straightforward to derive for multivariate kernel variants as well.

The proposed generalised spectral mixture (GSM) kernel for univariate inputs is
\begin{align}
k_{\text{GSM}}(x,x') = \sum_{i=1}^Q & w_i(x) w_i(x') \sqrt{ \frac{2 \ell_i(x) \ell_i(x') }{\ell_i(x)^2 + \ell_i(x')^2} } \exp\left(- \frac{(x-x')^2}{\ell_i(x)^2 + \ell_i(x')^2} \right) \\
 & \times \cos\left(2 \pi (\mu_i(x) x - \mu_i(x') x')\right)
\end{align}
with Gaussian process functions $w_i(x), \mu_i(x), \ell_i(x)$. The Spectral Mixture (SM) kernel \citep{wilson2013} is given by
\begin{align}
k_{\text{SM}}(x,x') &= \sum_{i=1}^Q w_i^2  \exp( -2 \pi^2 (x-x')^2 \sigma_i^2) \cos( 2 \pi \mu_i (x-x')) \,,
\end{align}
where the parameters are the weights $w_i$, mean frequencies $\mu_i$ and variances $\sigma_i^2$. Now if we assign the following constant functions for the GSM kernel to match the parameters of the SM kernel on the right-hand side,
\begin{align}
w_i(x) \equiv w_i, \qquad \mu_i(x) \equiv \mu_i, \qquad \ell_i(x) \equiv \frac{1}{2 \pi \sigma_i},
\end{align}
we retrieve the SM kernel
\begin{align}
k_{\text{GSM}}(x,x') &= \sum_{i=1}^Q w_i^2 \sqrt{\frac{2/(2\pi\sigma_i)^2}{2/(2\pi\sigma_i)^2}}\exp\left(- \frac{(x-x')^2}{2 (1 / (2 \pi \sigma_i))^2} \right) \cos(2 \pi (\mu_i x - \mu_i x')) \\
 &= \sum_{i=1}^Q w_i^2 \exp\left(- 2 \pi^2 \sigma_i^2 (x-x')^2 \right) \cos(2 \pi \mu_i (x - x'))  = k_{\text{SM}}(x,x').
\end{align}
This indicates that the GSM kernel can reproduce any kernel that is reproducable by the SM kernel, which is known to be a highly flexible kernel \citep{wilson2013,wilson2015kissgp}. In practise we can simulate stationary kernels with the GP-GSM by setting the spectral function kernels $k_w, k_\mu, k_\ell$ to enforce very smooth, or in practise constant, functions.

\subsection{Relation to other Non-Stationary Kernels}

\citet{heinonen2016} proposed a fully non-stationary version of the Gaussian kernel, formulated as
\begin{align}
    k(x, x') = \s(x)\s(x')\sqrt{\frac{2\ell(x)\ell(x')}{\ell(x)^2+\ell(x')^2}}
                \exp\left(-\frac{(x - x')^2}{\ell(x)^2+\ell(x')^2}\right) .
\end{align}

We obtain the non-stationary Gaussian kernel of the adaptive GP \citep{heinonen2016} by using one component ($Q=1$) mixture and setting the frequency function to a constant $\mu(x) \equiv 0$. The adaptive GP furthermore introduced a heteroscedastic noise model that was modelled as a GP as well, while in our GSM framework the noise is assumed to be constant across the input-space. Combining these approaches in a future work would be interesting.

\citet{paciorek2004} proposed a class of non-stationary kernels, formulated as
\begin{align}
    C^{\text{NS}}(\x_i, \x_j) &= |\Sigma_i|^{1/4}|\Sigma_j|^{1/4}| (\Sigma_i+\Sigma_j)/2|^{-1/2} R^\text{S}(\sqrt{Q_{ij}}) \\
    Q_{ij} &= (\x_i - \x_j)^T ((\Sigma_i+\Sigma_j)/2)^{-1} (\x_i - \x_j)
\end{align}
where $R^\text{S}(\cdot)$ is a stationary positive definite kernel, specifically considered to be a Matern as well as a Gaussian covariance. In the univariate case the $\Sigma_i$ and $\Sigma_j$ are just input-dependent scalar-valued functions, which are given Gaussian process priors. The adaptive GP \citep{heinonen2016} extended this to the noise and signal variance terms as well. The GSM kernel further extends these kernels to a mixture kernel with a cosine term that came from the spectral mixture formulation.

\section{Inference}

We apply Gaussian processes in the regression framework, and assume a Gaussian likelihood over $N$ data points $(\x_j,y_j)_{j=1}^{N}$ with all outputs collected into a vector $\y \in \R^{N}$,
\begin{align}
y_j &= f(\x_j) + \varepsilon_j, \qquad \varepsilon_j \sim \N(0, \sigma_n^2) \label{eq:likelihood} \\
 f(\x) &\sim \GP(0, k(\x,\x' \mid \bt)),
\end{align}
where \eqref{eq:likelihood} implies a Gaussian likelihood
\begin{align}
    p(\y \mid \f) = \N(\y \mid \f, \sigma_n^2 \mat I) \, .
\end{align}
A standard approach for finding kernel parameters $\bt$ is to marginalize the function values $\f$,
and optimize the marginal log likelihood 
\begin{align}
    \log p(\y \mid \bt) = \log \E_{p(\f)} [p(\y \mid \f)]
\end{align}
using standard optimization techniques, such as gradient descent. This requires $\mathcal{O}(N^3)$ computation due to inverting $N \times N$ kernel matrix $\K$, making it unfeasible to apply GP's to large datasets.

In this paper we adopt a sparse stochastic variational approach to large-scale Gaussian processes \citep{hensman15scalable}, which has the main feature of being a very general way of approximating the GP without making strict assumptions on the kernel. There exists also approaches that are able to scale exact (or almost exact) inference to large datasets, when specific structures are present in the kernel matrix. Methods exploiting Kronecker structure \citep{wilson2014fast,wilson2015kissgp,saatcci2011scalable} in kernel matrices are most suitable for input dimensions 2--5, but not for one-dimensional data which we mainly consider in this paper, even though the proposed kernel is applicable in higher dimensions as well. Toeplitz algebra \citep{cunningham2008fast,wilson2015kissgp} enables fast inference but requires a regular grid where kernel is evaluated as well as requiring that the kernel is stationary, making the approach not suitable for the non-stationary kernels considered in this paper. State space methods \citep{nickisch18a,hartikainen2010kalman} as well as random Fourier features and sparse spectrum kernels \citep{hensman2018variational,lazaro2010sparse,gal2015improving} also are derived only for stationary kernels. Fourier features have been recently introduced for a class of non-stationary kernels as well \citep{ton2018spatial}.

In the following, we follow the development by \citet{hensman15scalable}. In sparse Gaussian processes the function values are decomposed by adding new extra variables $\u$ giving the joint distribution as
\begin{align}
    p(\y, \f, \u \mid \bt) = p(\y \mid \f) p(\f \mid \u, \bt) p(\u \mid \bt) \label{eq:joint}
\end{align}
Here $\u$ denote the function values at \emph{inducing points} $\z$, and have the prior 
\begin{align}
    p(\u \mid \bt) = \N(\u \mid 0, \K_{\z\z}),
\end{align}
where $\K_{\z\z}$ denotes the kernel matrix computed at the inducing points. The kernel computation depends on the kernel hyperparameter values $\bt$, which for the GSM kernels include the parameterizations of the latent functions $\mu(x)$, $\ell(x)$ and $w(x)$, specifics for the case of GP-GSM are discussed at the end of this section.
The posterior $p(\u \mid \y)$ is approximated by $q(\u) = \N(\u \mid \m, \S)$, where $\m$ and $\S$ are the variational parameters. 

The following inequality applies for the joint distribution \eqref{eq:joint},
\begin{align}
    \log p(\y \mid \u) \geq \E_{p(\f \mid \u)} [\log p(\y \mid \f)] .
\end{align}
Additionally we have the standard variational bound
\begin{align}
    \log p(\y \mid \bt) \geq \E_{q(\u)} [\log p(\y \mid \u, \bt)] - \KL{q(\u)}{p(\u \mid \bt)}
\end{align}
using the approximate posterior $q(\u)$.
The evidence lower bound (ELBO) for the sparse variational Gaussian process is constructed by applying the two separate lower bounds to the marginal likelihood.
\begin{align}
    \log p(\y \mid \bt) &\geq \E_{q(\u)} [\log p(\y \mid \u, \bt)] - \KL{q(\u)}{p(\u \mid \bt)} \\
    &\geq \E_{q(\u)} [\E_{p(\f \mid \u, \bt)} [\log p(\y \mid \f)]] - \KL{q(\u)}{p(\u \mid \bt)} \label{eq:svgp}
\end{align}
The ELBO is then jointly maximized with respect to the variational parameters $\m$ and $\S$, inducing points $\z$ as well as any hyperparameters $\bt$ belonging to the kernel functions and possible hyperparameters part of the likelihood (i.e. the noise variance in case of regression). 
The covariance $\S$ is decomposed into $\S = \LL \LL^T$, where $\LL$ is constrained to a lower triangular matrix whose elements can be freely optimized.
GPflow \citep{GPflow2017} implements the variational bound \eqref{eq:svgp} for several non-conjugate likelihoods and link functions exactly (including Gaussian as well as Bernoulli with a probit link), and falls back to numerical integration for the ones that not tractable.

For the GP-GSM kernel that has parameters defined as Gaussian processes, we need to apply the inducing point approach also to the parameters of the kernel. 
Full variational treatment of the latent function values $\mu(x)$, $\ell(x)$ and $w(x)$ is not tractable.
In this work we assume that the kernel parameters and the outputs share the same inducing point locations $\z$ as the function values $\u$. 
Now, instead of having a full variational distribution for the latent functions, we simply find a point estimate for them.
When evaluating the GSM kernel for any new point $x$ (both during training and test time), we compute the latent function values as
\begin{align}
    \mu(x) &= \K_{x\z} \K_{\z\z}^{-1} \u_{\bmu} \\
    \ell(x) &= \K_{x\z} \K_{\z\z}^{-1} \u_{\vec\ell} \\
    w(x) &= \K_{x\z} \K_{\z\z}^{-1} \u_{\w} \, ,
\end{align}
where $\vec U = \{\u_{\bmu}, \u_{\vec\ell}, \u_{\w}\}$ are the values of the parameter functions at the inducing points $\z$, similarly to the output function values $\m$ in $q(\u \mid \m, \S)$.
The set of parameters that are to be learned for GP-GSM are thus $\bt = \{ \m, \S, \vec U, \sigma_n^2 \}$.

The RBF, SM and Neural-GSM kernels on the other hand do not need any special consideration within the variational inference framework, as we are just seeking point estimates of the kernel hyperparameters with respect to the ELBO. For Neural-GSM, the set of parameters to be learned is thus $\bt = \{ \m, \S, \W, \bb, \sigma_n^2 \}$, where $\W$ and $\bb$ are the weights and biases of the neural networks.

\section{Experiments}\label{sec:experiments}

We apply our proposed kernel first on simple simulated time series, then on two different solar time series (sunspot number and solar irradiance) and lastly on motion capture data. We compare our methods to the stationary spectral mixture (SM) \citep{wilson2013}, and the standard RBF kernels. Summary of the results is reported in Table~\ref{tab:results}.

We implemented all three spectral kernels discussed above using Tensorflow with the GPflow library \citep{GPflow2017}. The RBF kernel is implemented in GPflow. 
A Python package implementing the different spectral mixture kernels is available on GitHub\footnote{\url{https://github.com/sremes/nssm-gp}}, along with the code to run the presented experiments.

With all methods, we sample several random initializations for the kernel hyperparameters and select the best initialization which we optimize further. This is done to avoid some possibly very bad or pathological initializations. 
To determine the best settings for kernels, we run them with multiple settings, specifically trying the combinations of number of mixture components $Q \in \{1, 2, 3\}$, learning rates 0.01 and 0.001, and batch sizes 64 and 128. In all experiments we use $M = 100$ inducing points. With Neural-GSM, we use two hidden layers with 32 units each and L2 regularization for the weights.
We run all experiments on a shared computation cluster with each job reserved 8 CPU cores and 16 gigabytes of memory.

\begin{table}[]
    \caption{Likelihoods and absolute and square errors on the regression benchmarks, solar datasets and the motion capture dataset using the various kernels. Best result on each dataset in bold. The size ($N$) and the dimensionality ($D$) of the datasets are indicated below the names of the datasets.}
    \label{tab:results}
    \centering
    
    \begin{tabular}{lllll}  
    \toprule
    Dataset    & Kernel       &  $\log p(\y)$      & Mean absolute error & Mean square error \\
    
    \midrule
    Power      & Neural-GSM   & \textbf{-0.0204} $\pm$ 0.006  & \textbf{0.177} $\pm$ 0.001 & \textbf{0.0607} $\pm$ 0.0006 \\
    $N=9568$   & GP-GSM       & -0.0638 $\pm$ 0.02   & 0.183 $\pm$ 0.005 & 0.0690 $\pm$ 0.008  \\
    $D=4$      & SM           & -0.0667 $\pm$ 0.01   & 0.190 $\pm$ 0.003 & 0.0668  $\pm$ 0.001  \\
    ~          & RBF          & -0.0452 $\pm$ 0.0005 & 0.183 $\pm$ 0.0004& 0.0637 $\pm$ 0.00007\\
    \midrule
    Protein    & Neural-GSM   & \textbf{-1.01} $\pm$ 0.005 & \textbf{0.482} $\pm$ 0.004  & \textbf{0.438} $\pm$ 0.004 \\
    $N=45730$  & GP-GSM       & -1.40 $\pm$ 0.02  & 0.865 $\pm$ 0.02   & 0.965 $\pm$ 0.03  \\
    $D=9$      & SM           & -1.08 $\pm$ 0.004 & 0.555 $\pm$ 0.003  & 0.510 $\pm$ 0.005 \\
    ~          & RBF          & -1.07 $\pm$ 0.001 & 0.546 $\pm$ 0.0004 & 0.499 $\pm$ 0.001 \\
    \midrule
    Irradiance & Neural-GSM   &  0.0756 $\pm$ 0.3  & 0.168 $\pm$ 0.03 & \textbf{0.0459} $\pm$ 0.02 \\
    $N=391$    & GP-GSM       & -0.254 $\pm$ 0.9 &  0.181 $\pm$ 0.1 & 0.0918 $\pm$ 0.1 \\
    $D=1$      & SM           &  \textbf{0.162} $\pm$ 0.4 & \textbf{0.147} $\pm$ 0.05 & 0.0492 $\pm$ 0.03 \\    
    ~          & RBF          & -0.730 $\pm$ 0.2 & 0.337 $\pm$ 0.05 & 0.179 $\pm$ 0.06 \\
    \midrule
    Sunspots   & Neural-GSM   & \textbf{-0.527} $\pm$ 0.01  & \textbf{0.292} $\pm$ 0.006  & \textbf{0.163} $\pm$ 0.004\\
    $N=1599$   & GP-GSM       & \textbf{-0.526} $\pm$ 0.02  & 0.295 $\pm$ 0.006  & 0.167 $\pm$ 0.005 \\
    $D=1$      & SM           & -0.540 $\pm$ 0.02  & 0.296 $\pm$ 0.007  & 0.166 $\pm$ 0.006 \\    
    ~          & RBF          & -1.38  $\pm$ 0.002 & 0.802 $\pm$ 0.002  & 0.940 $\pm$ 0.003 \\
    \midrule
    Motion     & Neural-GSM   & 0.512 $\pm$ 0.02 & 0.0963 $\pm$ 0.004 & 0.0193 $\pm$ 0.001 \\
    $N=6220$   & GP-GSM       & \textbf{0.750} $\pm$ 0.02 & \textbf{0.0739} $\pm$ 0.003 & \textbf{0.0128} $\pm$ 0.0005\\
    $D=1$      & SM           & 0.389 $\pm$ 0.07 & 0.106 $\pm$ 0.007 & 0.0249 $\pm$ 0.004 \\  
    ~          & RBF          & 0.129 $\pm$ 0.04 & 0.133 $\pm$ 0.003 & 0.0416 $\pm$ 0.004\\
    \end{tabular}
\end{table}

\subsection{Time-Series Datasets}

\subsubsection{Solar Datasets}

\begin{figure}
    \centering
    \includegraphics[width=0.32\textwidth]{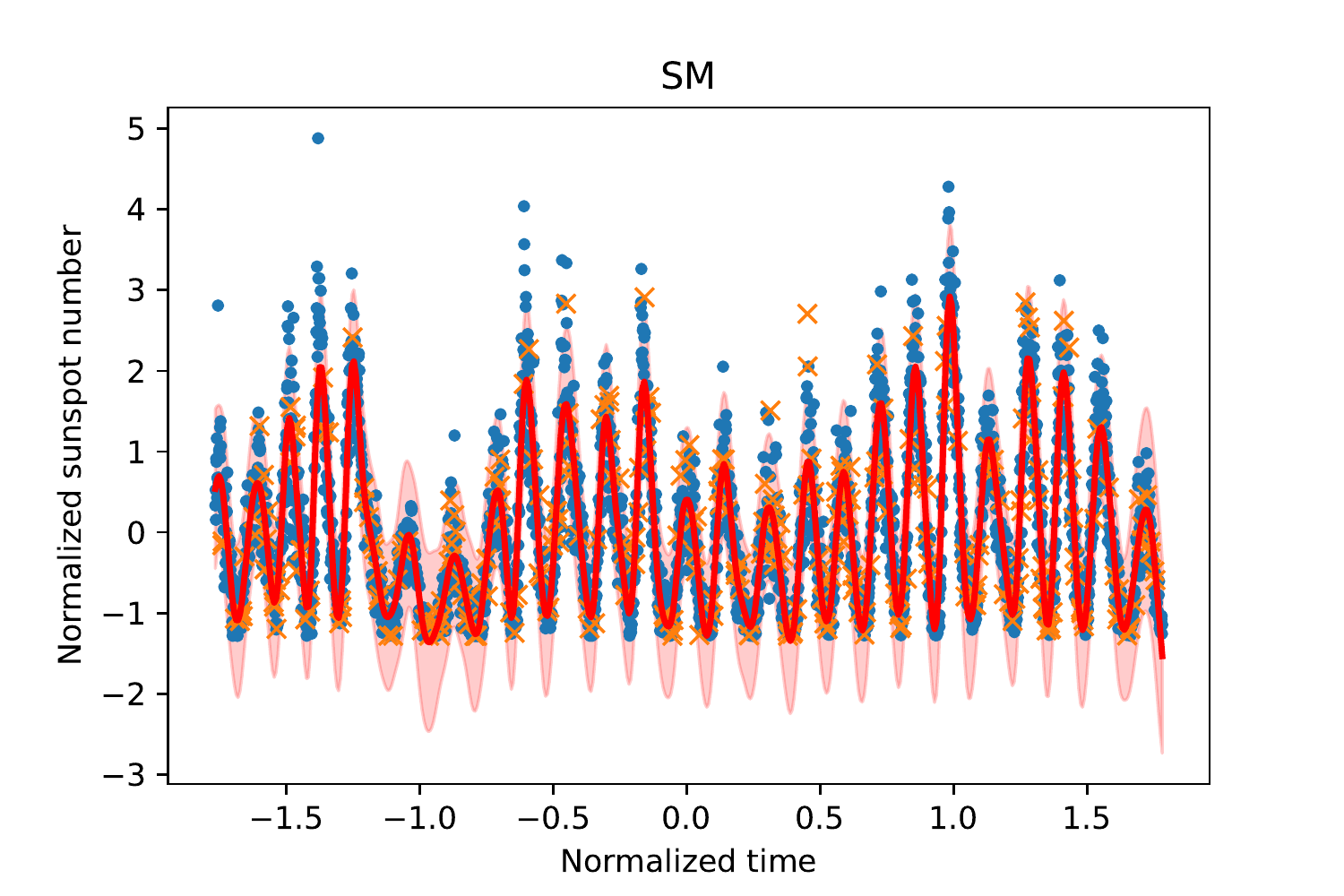}
    \includegraphics[width=0.32\textwidth]{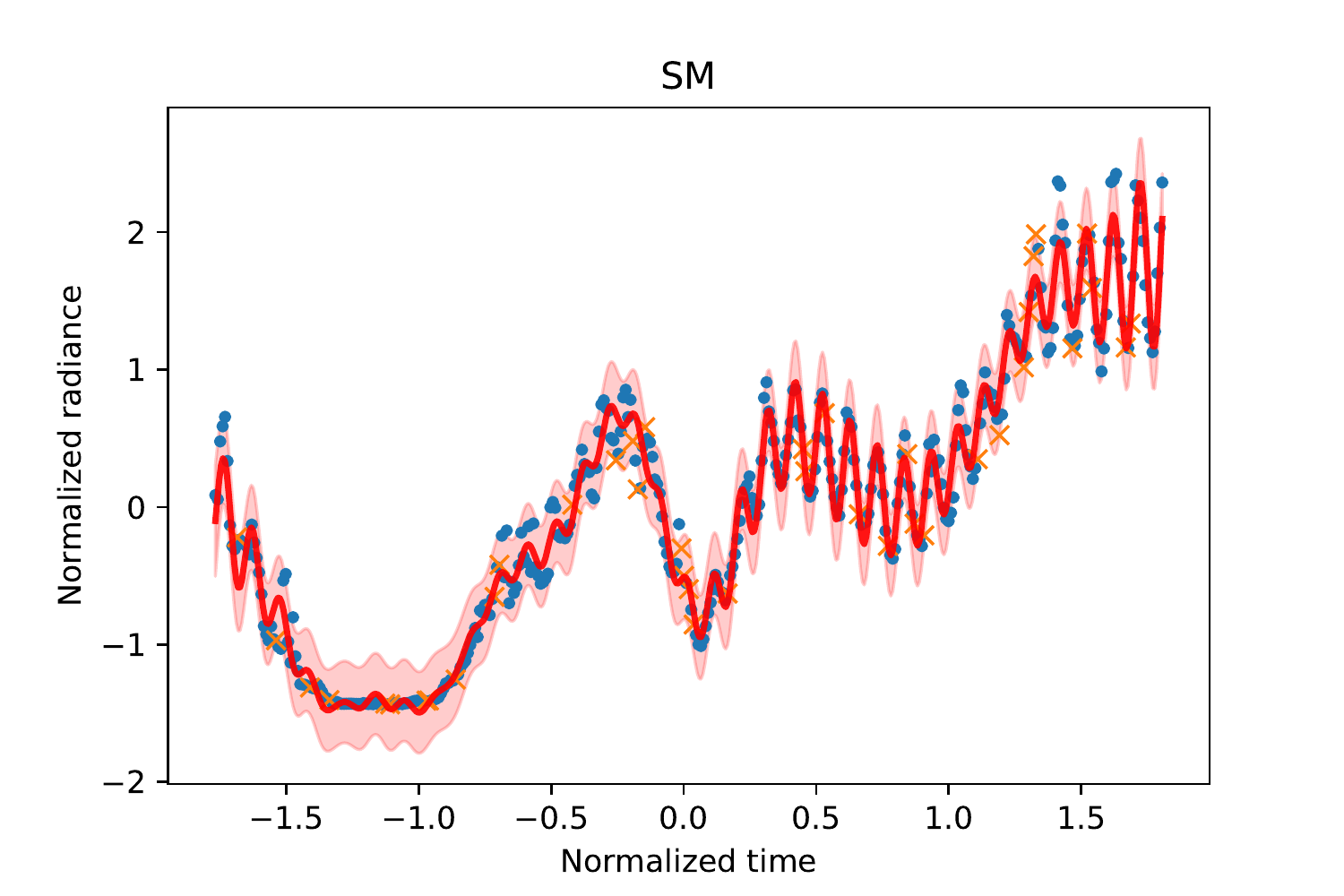}
    \includegraphics[width=0.32\textwidth]{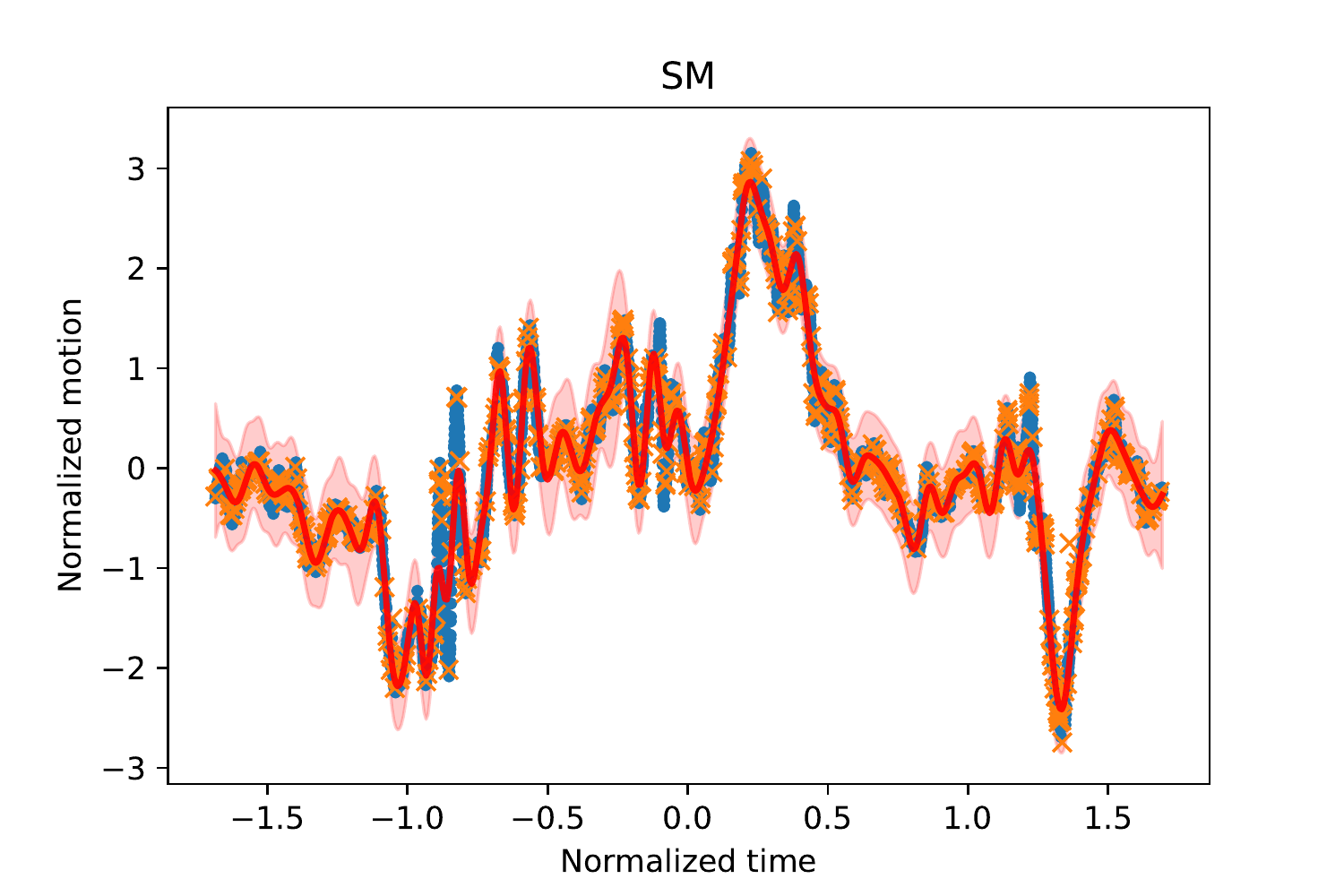}
    
    \includegraphics[width=0.32\textwidth]{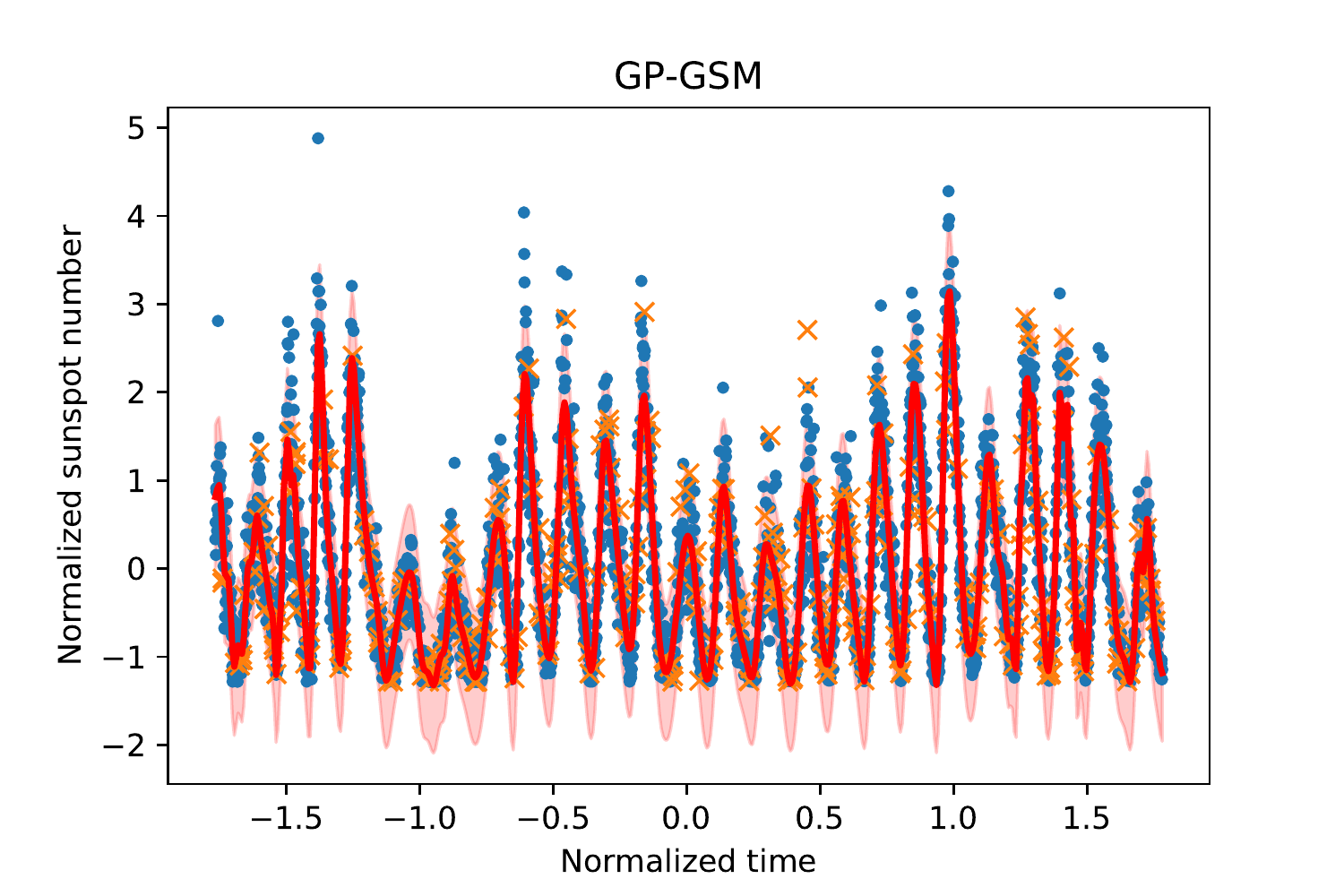}
    \includegraphics[width=0.32\textwidth]{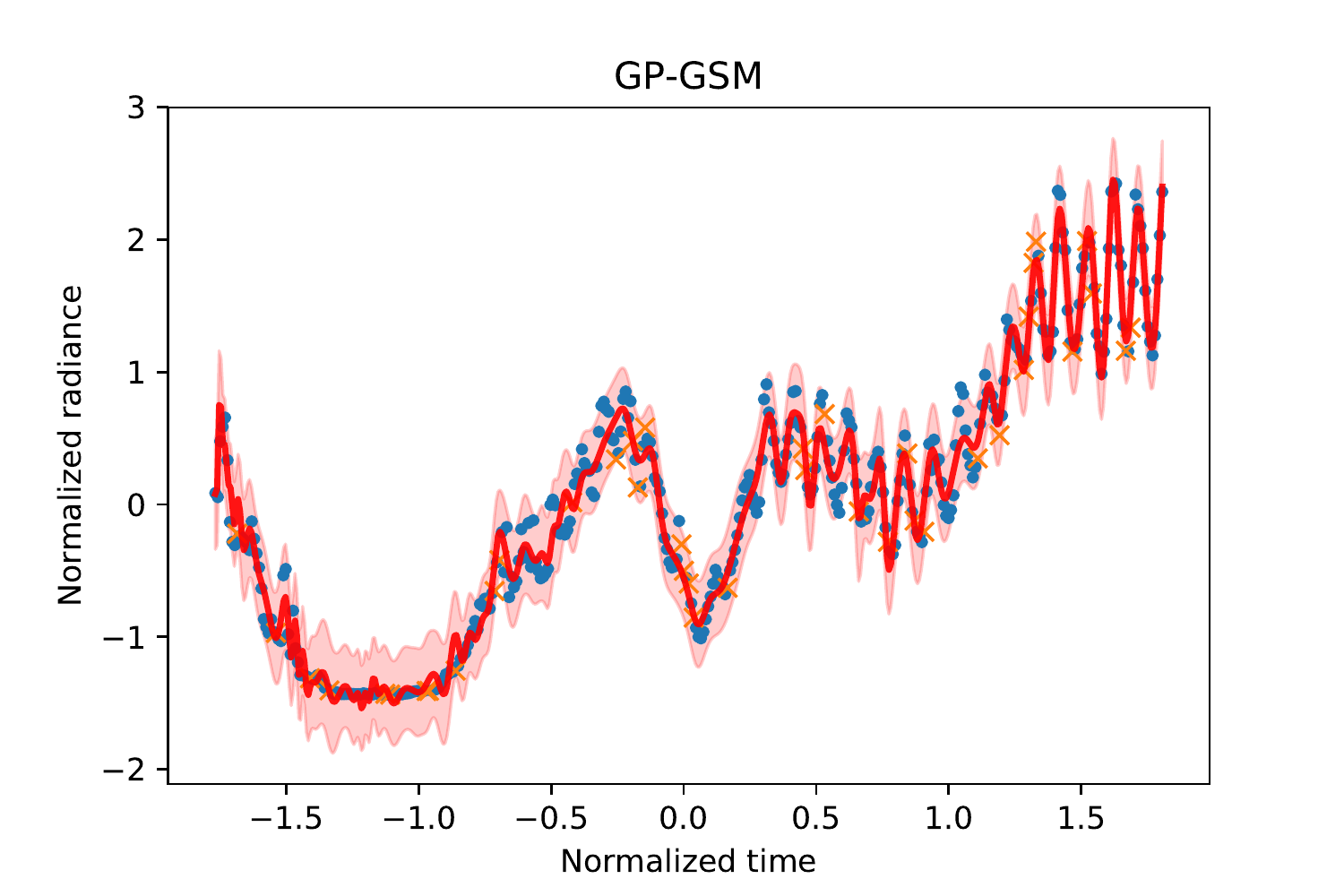}
    \includegraphics[width=0.32\textwidth]{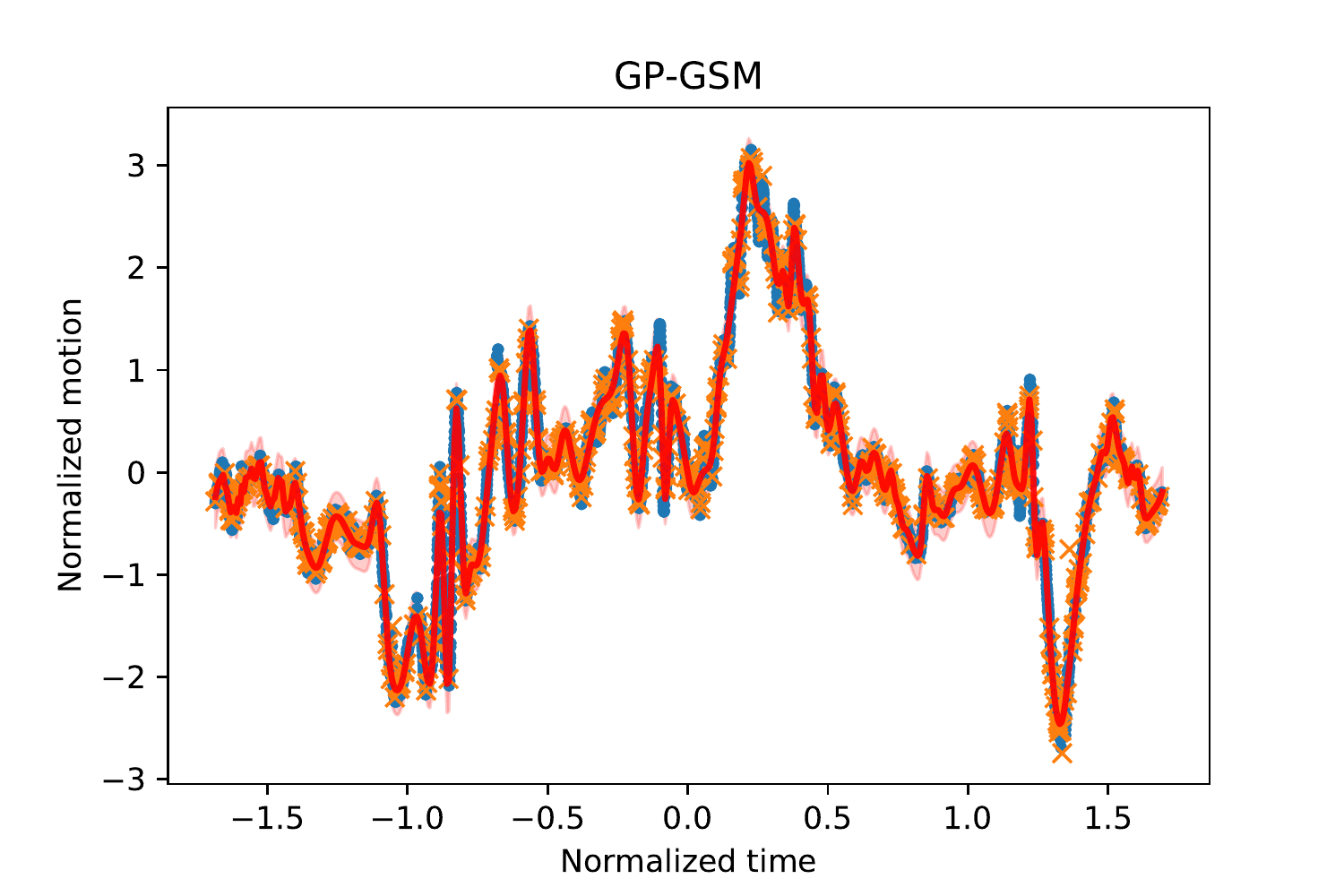}
    
    \includegraphics[width=0.32\textwidth]{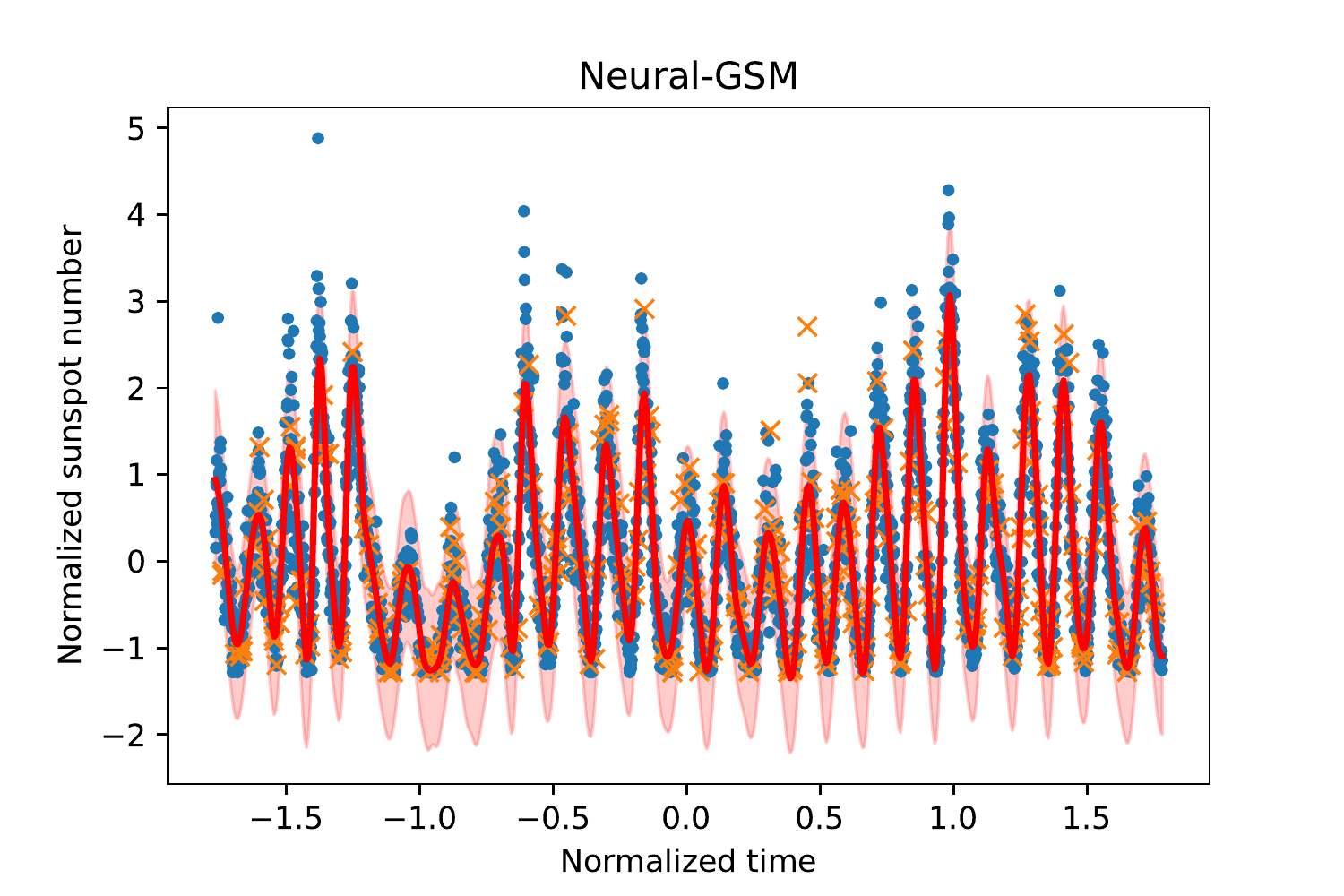}
    \includegraphics[width=0.32\textwidth]{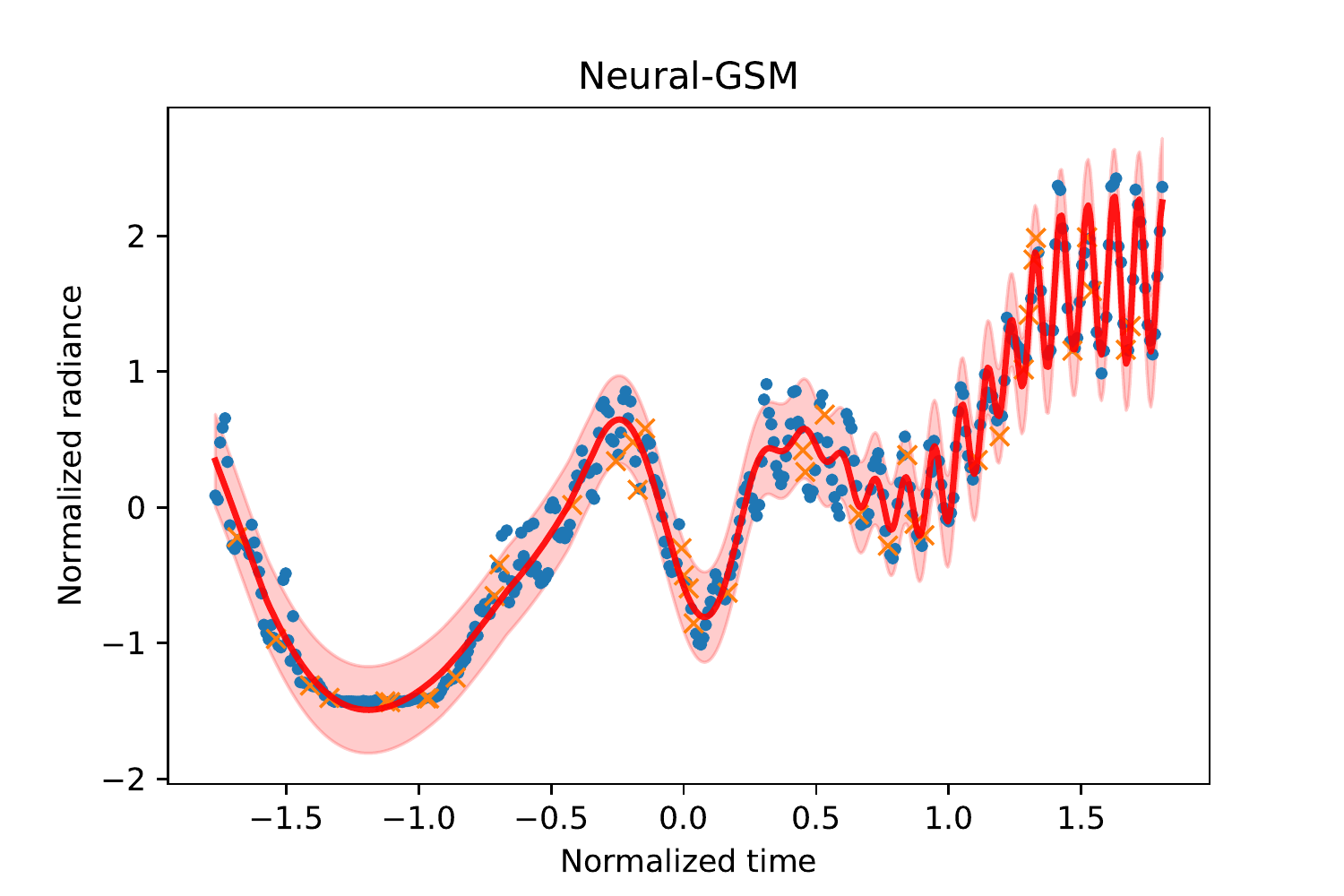}
    \includegraphics[width=0.32\textwidth]{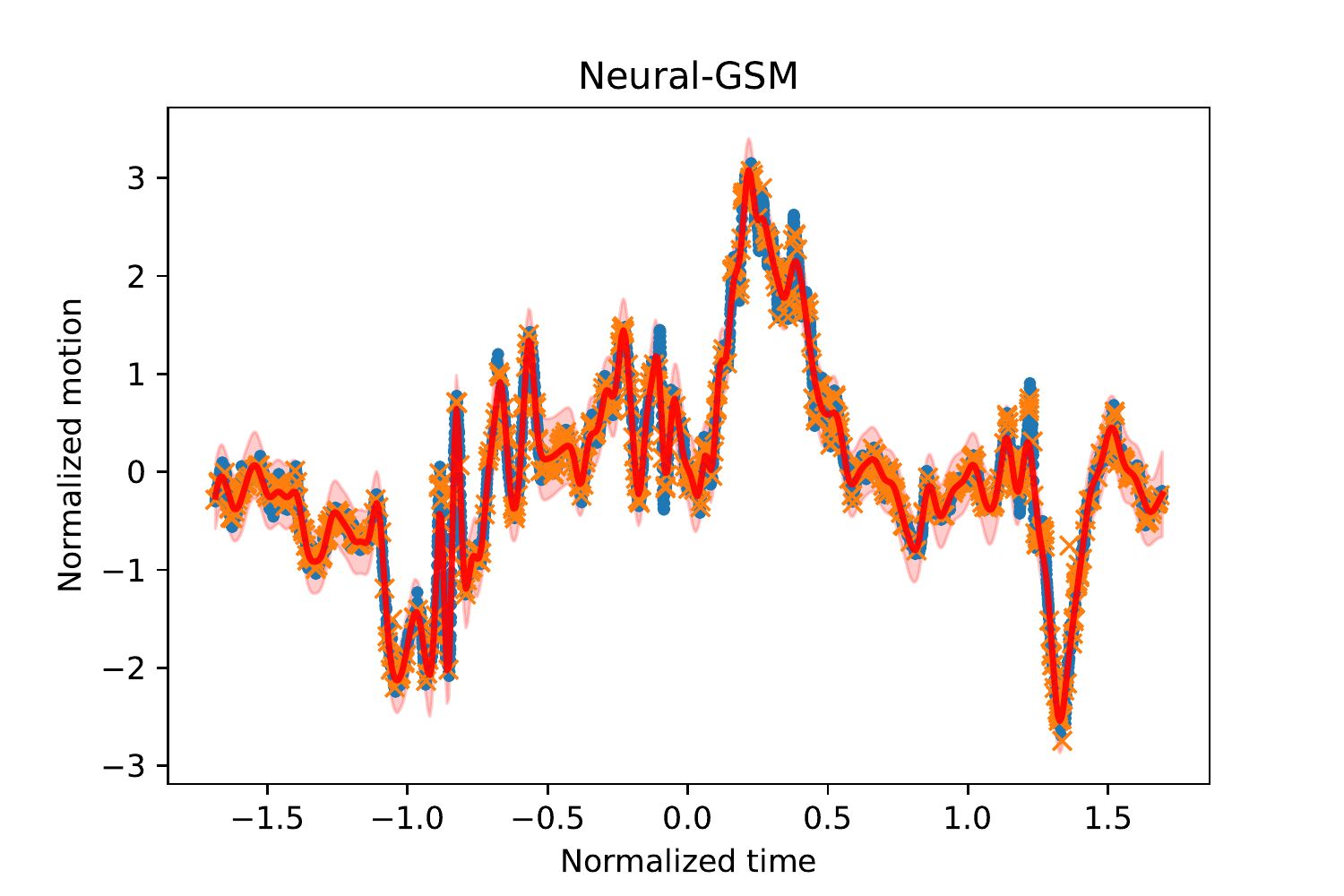}
    \caption{GP posteriors for the solar datasets, sunspots (left column) and irradiance (mid-column), and the CMU motion capture (right column) with the three spectral kernel variants. Training data is indicated by blue dots, test data for which we report accuracy measures as orange crosses and the GP posterior with solid line for the mean and confidence interval as a shaded area.}
    \label{fig:timeseries}
\end{figure}

We compare the kernels on two different solar activity datasets. 
The solar irradiance dataset \citep{lean2004} consists of annually reconstructed time series of the Sun's spectral irradiance between years 1610 and 2000.
The sunspots dataset \citep{sidc} includes monthly mean total sunspot number, defined simply as the arithmetic mean of daily sunspot number within the month. The timespan measured is from February 1749 until August 2018.
Resulting GP posteriors on all datasets with the different kernels are depicted in Figure~\ref{fig:timeseries}. The RBF kernel is not shown as the bad fit by the kernel obscures the plots. The marginal log likelihoods, the mean absolute errors and the mean square errors are summarised in Table~\ref{tab:results}.

None of the kernels are perfectly able to fit to the irradiance dataset, which contains time periods with no periodic behaviour as well as periods with strong periodicity. The optimization of the kernel hyperparameters is unstable, as the lower bound likely contains several local optima that correspond to either smooth or periodic solutions, and even the non-stationary kernels likely prefer either one of the two possible solutions. The differences in performance between the kernels are not significant given the standard deviations.

On the sunspots data, Neural-GSM provides the best fit.  The Neural variant of the GSM produces slightly better results, with the GP-GSM being second in performance on this data. The SM kernel seems to fit the periodicity well, but the amplitude is too low at the end of the time series, where the periodicity is strongest. The non-stationary versions fit that region well, but infer that there is less periocity in other parts of the time series.

\subsubsection{Motion capture}

Motion capture data\footnote{Available from CMU Graphics Lab Motion Capture Database \url{http://mocap.cs.cmu.edu/}.} consists of time series of many motion sensors recorded during a movement of a subject, performing movement such as walking, running or jumping. The CMU motion capture records the movements at 120 Hz. We took one of the longer captures consisting of variety of different movements (subject 56, trial \#3), which has motion such as wiping windows, yawning, stretching, angry grabbing, smashing against wall and skipping. 
The capture is 52 seconds long, having 6220 frames. 
The data is preprocessed with normalizing all features and taking the first PCA component as the representative motion time series.

The data and the GP posteriors with the different kernels are shown in Figure~\ref{fig:timeseries}, and performance measures are presented in Table~\ref{tab:results}.
On this data the two GSM variants give the best performance
The stationary SM kernel ends up with too wide confidence intervals for the predictions, and misses some of the sharper peaks, that the non-stationary GSM kernels are able to fit.

\begin{figure}
    \centering
    \includegraphics[width=0.7\textwidth]{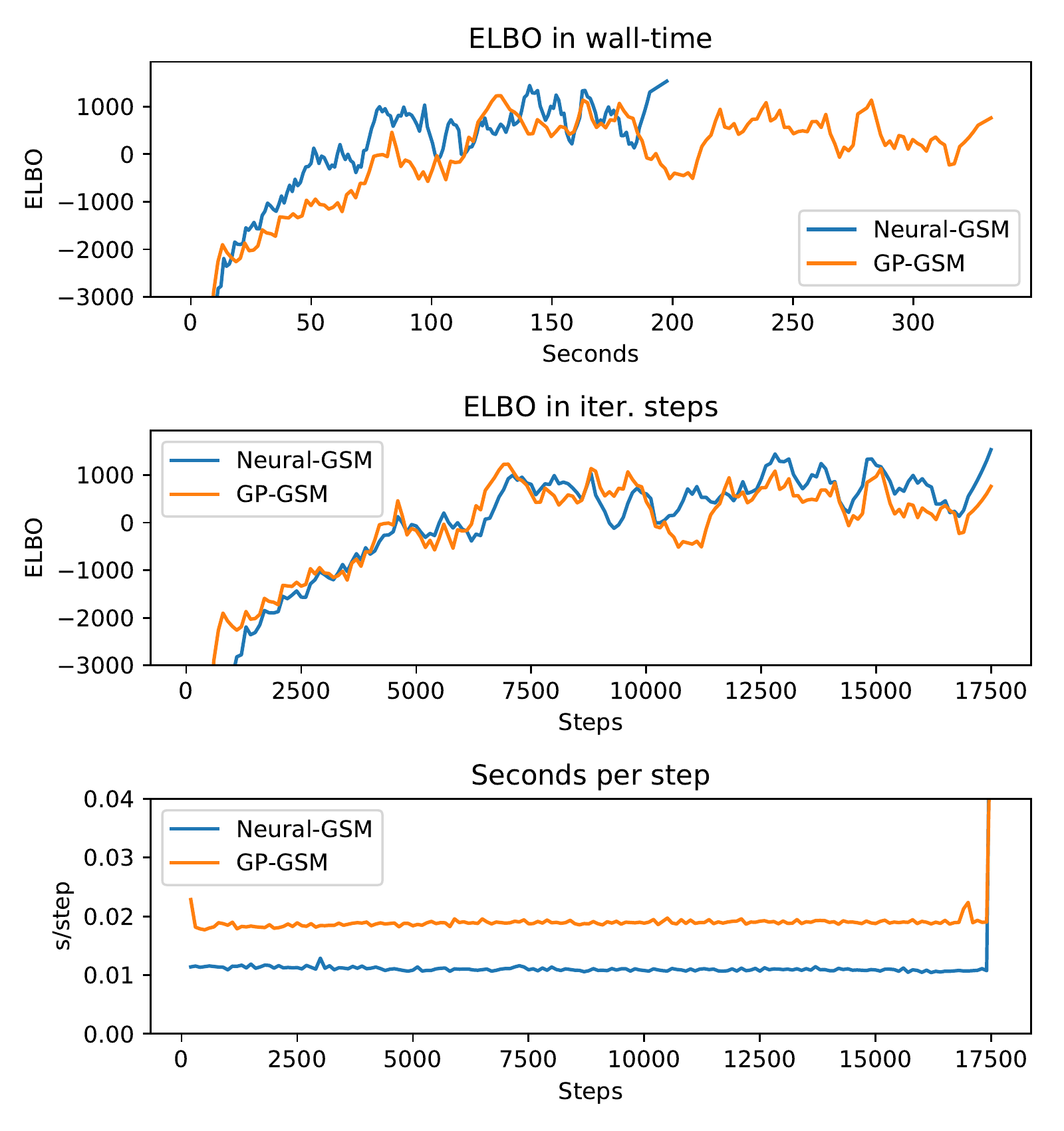}
    \caption{Comparison of running times on the CMU motion capture dataset with Neural-GSM and GP-GSM. In this example, Neural-GSM runs almost twice as fast with respect to wall-time, while requiring approximately the same number of iterations to converge as GP-GSM. The ELBO curves are slightly smoothed with a Savitzky–Golay filter. Note that on the top-most figure, both kernels are run the same number of iterations (approximately 17500).}
    \label{fig:runtimes}
\end{figure}

For this experiment, we also studied the running times between the proposed Neural-GSM and the GP-GSM. In Figure~\ref{fig:runtimes}, we plot the ELBO during learning as a function of optimization steps as well as as a function of wall-time. Both kernels appear to improve approximately at the same rate with respect to optimization steps, but the Neural-GSM is almost twice as fast as GP-GSM to compute per iteration.

\subsection{Regression benchmarks: Power and Protein}

Furthermore, we compare the kernels on two standard UCI\footnote{\url{https://archive.ics.uci.edu/ml/index.php}} repository regression benchmarks: the Power and Protein datasets. These are among the lowest dimensional datasets available on the repository, making them still suitable for spectral kernels that have many dimension specific hyperparameters. Results are show in Table~\ref{tab:results}. Neural-GSM performs best on both datasets, while SM, GP-GSM and RBF are more even compared to each other.
GP-GSM appeared to be somewhat more unstable in converging to reasonable solutions, so we ran it a few more times than other models. The instability in GP-GSM is also evident in the standard deviation estimates for the likelihood and errors among the top models.

\section{Discussion}\label{sec:discussion}

This paper introduced a new formulation for the generalized spectral mixture kernel, where the parameters were modelled with neural networks instead of Gaussian processes, as was done by \citet{remes2017}. Furthermore we use more scalable inference techniques, specifically sparse stochastic variational GP's, to enable our method to scale to several thousands of data points in a single dimension, as opposed to the Kronecker variant in \citet{remes2017} for GP-GSM. We also implemented the GP-GSM within the stochastic variational framework.

We show that the Neural-GSM performs better or as well as the comparison methods in most of the experiments.
In the time-series experiments, the RBF kernel often had too wide length-scales resulting in underfitting the data. This is likely caused by the sparse GP approximation being unable interpolate from the inducing points to the observed inputs due to the fact the RBF kernel cannot learn any longer range structures (e.g. periodicities), which the spectral mixtures are capable of. On the general regression benchmarks, Protein and Power, the Neural-GSM kernel performs the best.

The proposed Neural-GSM kernel is also approximately twice as fast in terms of computation needed per iteration within the variational inference framework, compared with the previously proposed GP-parameterized GSM kernel.

\subsection*{Acknowledgements}
This work has been supported by the Academy of Finland (grants 294238 and 292334, and Finnish Centre of Excellence in Computational Inference Research COIN). 
We also acknowledge the computational resources provided by the Aalto Science-IT project.

\bibliographystyle{spbasic}
\bibliography{refs}

\begin{thebibliography}{42}
\providecommand{\natexlab}[1]{#1}
\providecommand{\url}[1]{{#1}}
\providecommand{\urlprefix}{URL }
\expandafter\ifx\csname urlstyle\endcsname\relax
  \providecommand{\doi}[1]{DOI~\discretionary{}{}{}#1}\else
  \providecommand{\doi}{DOI~\discretionary{}{}{}\begingroup
  \urlstyle{rm}\Url}\fi
\providecommand{\eprint}[2][]{\url{#2}}

\bibitem[{Cunningham et~al.(2008)Cunningham, Shenoy, and
  Sahani}]{cunningham2008fast}
Cunningham JP, Shenoy KV, Sahani M (2008) Fast {G}aussian process methods for
  point process intensity estimation. In: ICML, pp 192--199

\bibitem[{Gal and Turner(2015)}]{gal2015improving}
Gal Y, Turner R (2015) Improving the gaussian process sparse spectrum
  approximation by representing uncertainty in frequency inputs. In:
  International Conference on Machine Learning, pp 655--664

\bibitem[{Genton(2001)}]{genton2001}
Genton M (2001) Classes of kernels for machine learning: A statistics
  perspective. Journal of Machine Learning Research 2:299--312

\bibitem[{Gibbs(1997)}]{gibbs1997}
Gibbs M (1997) Bayesian {G}aussian processes for regression and classification.
  PhD thesis, University of Cambridge

\bibitem[{Goodfellow et~al.(2016)Goodfellow, Bengio, and
  Courville}]{goodfellow2016deep}
Goodfellow I, Bengio Y, Courville A (2016) Deep learning. MIT Press

\bibitem[{Gramacy and Lee(2008)}]{gramacy2008}
Gramacy R, Lee H (2008) Bayesian treed {G}aussian process models with an
  application to computer modeling. Journal of the American Statistical
  Association 103:1119--1130

\bibitem[{Grzegorczyk et~al.(2008)Grzegorczyk, Husmeier, Edwards, Ghazal, and
  Millar}]{Grzegorczyk2008}
Grzegorczyk M, Husmeier D, Edwards K, Ghazal P, Millar A (2008) Modelling
  non-stationary gene regulatory processes with a non-homogeneous {B}ayesian
  network and the allocation sampler. Bioinformatics 24:2071--2078

\bibitem[{Hartikainen and S{\"a}rkk{\"a}(2010)}]{hartikainen2010kalman}
Hartikainen J, S{\"a}rkk{\"a} S (2010) Kalman filtering and smoothing solutions
  to temporal gaussian process regression models. In: Machine Learning for
  Signal Processing (MLSP), 2010 IEEE International Workshop on, IEEE, pp
  379--384

\bibitem[{Heinonen et~al.(2016)Heinonen, Mannerström, Rousu, Kaski, and
  Lähdesmäki}]{heinonen2016}
Heinonen M, Mannerström H, Rousu J, Kaski S, Lähdesmäki H (2016)
  Non-stationary {G}aussian process regression with {H}amiltonian {M}onte
  {C}arlo. In: AISTATS, vol~51, pp 732--740

\bibitem[{Hensman et~al.(2015)Hensman, Matthews, and
  Ghahramani}]{hensman15scalable}
Hensman J, Matthews A, Ghahramani Z (2015) Scalable variational {G}aussian
  process classification. In: AISTATS, vol~38, pp 351--360

\bibitem[{Hensman et~al.(2018)Hensman, Durrande, and
  Solin}]{hensman2018variational}
Hensman J, Durrande N, Solin A (2018) Variational {F}ourier features for
  {G}aussian processes. Journal of Machine Learning Research 18(151):1--52

\bibitem[{Higdon et~al.(1999)Higdon, Swall, and Kern}]{higdon1999}
Higdon D, Swall J, Kern J (1999) Non-stationary spatial modeling. Bayesian
  statistics 6:761--768

\bibitem[{Huang(2008)}]{huang2008}
Huang N (2008) A review on {H}ilbert-{H}uang transform: Method and its
  applications to geophysical studies. Reviews of Geophysics 46

\bibitem[{Huang et~al.(1998)Huang, Zheng, Long, Wu, Shih, Zheng, Yen, Tung, and
  Liu}]{huang1998}
Huang N, Zheng S, Long S, Wu M, Shih H, Zheng Q, Yen NQ, Tung C, Liu H (1998)
  The empirical mode decomposition and the {H}ilbert spectrum for nonlinear and
  non-stationary time series analysis. In Proceedings of the Royal Society of
  London A: Mathematical, Physical and Engineering Sciences 454:903--995

\bibitem[{Kingma and Welling(2013)}]{kingma2013auto}
Kingma DP, Welling M (2013) Auto-encoding variational bayes. arXiv preprint
  arXiv:13126114

\bibitem[{Klambauer et~al.(2017)Klambauer, Unterthiner, Mayr, and
  Hochreiter}]{klambauer2017self}
Klambauer G, Unterthiner T, Mayr A, Hochreiter S (2017) Self-normalizing neural
  networks. In: NIPS, pp 971--980

\bibitem[{Kom~Samo and Roberts(2015)}]{komsamo2015}
Kom~Samo YL, Roberts S (2015) Generalized spectral kernels. Tech. rep.,
  University of Oxford, arXiv:1506.02236

\bibitem[{L{\'a}zaro-Gredilla et~al.(2010)L{\'a}zaro-Gredilla,
  Qui{\~n}onero-Candela, Rasmussen, and Figueiras-Vidal}]{lazaro2010sparse}
L{\'a}zaro-Gredilla M, Qui{\~n}onero-Candela J, Rasmussen CE, Figueiras-Vidal
  AR (2010) Sparse spectrum {G}aussian process regression. Journal of Machine
  Learning Research 11:1865--1881

\bibitem[{Lean(2004)}]{lean2004}
Lean J (2004) Solar irradiance reconstruction. IGBP PAGES/World Data Center for
  Paleoclimatology, Data Contribution Series \# 2004-035, NOAA/NGDC
  Paleoclimatology Program, Boulder CO, USA.

\bibitem[{Matthews et~al.(2017)Matthews, {van der Wilk}, Nickson, Fujii,
  {Boukouvalas}, {Le{\'o}n-Villagr{\'a}}, Ghahramani, and Hensman}]{GPflow2017}
Matthews AGdG, {van der Wilk} M, Nickson T, Fujii K, {Boukouvalas} A,
  {Le{\'o}n-Villagr{\'a}} P, Ghahramani Z, Hensman J (2017) {{GP}flow: A
  {G}aussian process library using {T}ensor{F}low}. Journal of Machine Learning
  Research 18(40):1--6, \urlprefix\url{http://jmlr.org/papers/v18/16-537.html}

\bibitem[{Nickisch et~al.(2018)Nickisch, Solin, and Grigorevskiy}]{nickisch18a}
Nickisch H, Solin A, Grigorevskiy A (2018) State space {G}aussian processes
  with non-{G}aussian likelihood. In: ICML, pp 3789--3798

\bibitem[{Paciorek and Schervish(2004)}]{paciorek2004}
Paciorek C, Schervish M (2004) Nonstationary covariance functions for
  {G}aussian process regression. In: NIPS, pp 273--280

\bibitem[{Paciorek and Schervish(2006)}]{paciorek2006}
Paciorek C, Schervish M (2006) Spatial modelling using a new class of
  nonstationary covariance functions. Environmetrics 17(5):483--506

\bibitem[{Rahimi and Recht(2008)}]{rahimi2008}
Rahimi A, Recht B (2008) Random features for large-scale kernel machines. In:
  NIPS, pp 1177--1184

\bibitem[{Rasmussen and Williams(2006)}]{rasmussen2006}
Rasmussen CE, Williams C (2006) Gaussian processes for machine learning. MIT
  Press

\bibitem[{Remes et~al.(2017)Remes, Heinonen, and Kaski}]{remes2017}
Remes S, Heinonen M, Kaski S (2017) Non-stationary spectral kernels. In: NIPS

\bibitem[{Rioul and Martin(1991)}]{rioul1991}
Rioul O, Martin V (1991) Wavelets and signal processing. IEEE signal processing
  magazine 8:14--38

\bibitem[{Robinson and Hartemink(2009)}]{robinson2009}
Robinson J, Hartemink A (2009) Non-stationary dynamic {B}ayesian networks. In:
  NIPS, pp 1369--1376

\bibitem[{Saat{\c{c}}i(2011)}]{saatcci2011scalable}
Saat{\c{c}}i Y (2011) Scalable inference for structured {G}aussian process
  models. PhD thesis, University of Cambridge

\bibitem[{Sampson and Guttorp(1992)}]{sampson1992}
Sampson P, Guttorp P (1992) Nonparametric estimation of nonstationary spatial
  covariance structure. Journal of the American Statistical Association 87

\bibitem[{{SILSO World Data Center}(1749-2018)}]{sidc}
{SILSO World Data Center} (1749-2018) The international sunspot number.
  International Sunspot Number Monthly Bulletin and online catalogue
  \url{http://www.sidc.be/silso/}

\bibitem[{Silverman(1957)}]{silverman1957}
Silverman R (1957) Locally stationary random processes. Information Theory, IRE
  Transactions on 3:182--187

\bibitem[{Sinha and Duchi(2016)}]{sinha2016}
Sinha A, Duchi J (2016) Learning lernels with random features. In: NIPS

\bibitem[{Snoek et~al.(2014)Snoek, Swersky, Zemel, and Adams}]{snoek2014}
Snoek J, Swersky K, Zemel R, Adams R (2014) Input warping for {B}ayesian
  optimization of non-stationary functions. In: ICML, vol~32, pp 1674--1682

\bibitem[{Tolvanen et~al.(2014)Tolvanen, Jyl{\"a}nki, and
  Vehtari}]{tolvanen2014}
Tolvanen V, Jyl{\"a}nki P, Vehtari A (2014) Expectation propagation for
  nonstationary heteroscedastic {G}aussian process regression. In: Machine
  Learning for Signal Processing (MLSP), 2014 IEEE International Workshop on,
  IEEE, pp 1--6

\bibitem[{Ton et~al.(2018)Ton, Flaxman, Sejdinovic, and Bhatt}]{ton2018spatial}
Ton JF, Flaxman S, Sejdinovic D, Bhatt S (2018) Spatial mapping with {G}aussian
  processes and nonstationary {F}ourier features. Spatial Statistics

\bibitem[{Wilson and Nickisch(2015)}]{wilson2015kissgp}
Wilson A, Nickisch H (2015) Kernel interpolation for scalable structured
  {G}aussian processes ({KISS}-{GP}). In: International Conference on Machine
  Learning, pp 1775--1784

\bibitem[{Wilson(2014)}]{wilson2014thesis}
Wilson AG (2014) Covariance kernels for fast automatic pattern discovery and
  extrapolation with {G}aussian processes. PhD thesis, University of Cambridge

\bibitem[{Wilson and Adams(2013)}]{wilson2013}
Wilson AG, Adams R (2013) Gaussian process kernels for pattern discovery and
  extrapolation. In: ICML

\bibitem[{Wilson et~al.(2014)Wilson, Gilboa, Nehorai, and
  Cunningham}]{wilson2014fast}
Wilson AG, Gilboa E, Nehorai A, Cunningham JP (2014) Fast kernel learning for
  multidimensional pattern extrapolation. In: NIPS, pp 3626--3634

\bibitem[{Wilson et~al.(2016)Wilson, Hu, Salakhutdinov, and
  Xing}]{wilson2016deep}
Wilson AG, Hu Z, Salakhutdinov R, Xing EP (2016) Deep kernel learning. In:
  Artificial Intelligence and Statistics, pp 370--378

\bibitem[{Yang et~al.(2015)Yang, Smola, Song, and Wilson}]{yang2015}
Yang Z, Smola A, Song L, Wilson A (2015) A la carte: Learning fast kernels. In:
  AISTATS

\end{thebibliography}

\end{document}